\pdfoutput=1
\documentclass[conference]{IEEEtran}
\usepackage{xcolor,pifont}
\newcommand*\colourcheck[1]{%
  \expandafter\newcommand\csname #1check\endcsname{\textcolor{#1}{\ding{52}}}%
}
\colourcheck{blue}
\colourcheck{green}

\ifCLASSINFOpdf
  \usepackage[pdftex]{graphicx}
\else
\fi

\usepackage{stfloats}
\usepackage{url}

  \usepackage[frozencache=true,cachedir=mc]{minted}

\begin{document}
%
\title{Green Tsetlin \\ Redefining Efficiency in Tsetlin Machine Frameworks}

\author{
\IEEEauthorblockN{Sondre Glimsdal}
\IEEEauthorblockA{Forsta Laboratories\\
Center for AI Research\\
University of Agder\\
Grimstad Norway\\
sondre.glimsdal@forsta.com \\
0000-0001-9735-4064}
\and
\IEEEauthorblockN{Sebastian Østby}
\IEEEauthorblockA{Forsta Laboratories\\
Center for AI Research\\
University of Agder\\
Grimstad Norway\\
sebastiaos@uia.no \\
0000-0002-1159-2430}
\and
\IEEEauthorblockN{Tobias M. Brambo}
\IEEEauthorblockA{Forsta Laboratories\\
Center for AI Research\\
University of Agder\\
Grimstad Norway\\
tobiasmb@uia.no \\
0000-0002-7838-0467}
\and
\IEEEauthorblockN{Eirik M. Vinje}
\IEEEauthorblockA{Forsta Laboratories\\
Center for AI Research\\
University of Agder\\
Grimstad Norway\\
eirikmv@uia.no}
}


%


\maketitle

\begin{abstract}
Green Tsetlin (GT) is a Tsetlin Machine (TM) framework developed to solve real-world problems using TMs. Several frameworks already exist that provide access to TM implementations. However, these either lack features or have a research-first focus. GT is an easy-to-use framework that aims to lower the complexity and provide a production-ready TM implementation that is great for experienced practitioners and beginners.

To this end, GT establishes a clear separation between training and inference. A C++ backend with a Python interface provides competitive training and inference performance, with the option of running in pure Python. It also integrates support for critical components such as exporting trained models, hyper-parameter search, and cross-validation out-of-the-box.
\end{abstract}


%
\IEEEpeerreviewmaketitle

\section{Introduction}
Numerous implementations of the Tsetlin Machine \cite{granmo2021tsetlin} have been found since its inception.
These implementations are either a one-shot implementation for a particular research project or
an extension to an existing research code base. While this approach is acceptable for research, to
further the growth of TM, a framework that instead is focused on \emph{using} TMs is needed.

Green Tsetlin (GT) is a TM framework designed from the ground up to
be used for solving explainable ML tasks from a practical standpoint, including microcontrollers such
as the Arduino and Rasberry Pi. The GT framework provides faithful implementations of the Coaleased Tsetlin Machine \cite{glimsdal2021coalesced},
Sparse Tsetlin Machine\cite{ostby2024sparse}. That is, unless marked, GT tries to avoid implementation-specific
shortcuts or optimizations not covered in the papers \footnote{One such example is that pyTsetlinMachine only 
calculates $s$ per literal once per example, not per clause as indicated in \cite{granmo2021tsetlin}.}.
The result is a framework that can efficiently run anything from small-scale TMs to large-scale TMs. 

\section{Design Principles}
The design principles of GT are based on a collection of experiences from other GT frameworks.
These can be summarized in the following four principles:
\begin{itemize}
    \item{
        \textbf{Practical} \\
        GT does not strive to implement all variations and variants of the TM.
        Instead, it focuses on providing a high-quality implementation of a selected subset of TM variants and the utilities to perform high-quality work.
        These features should cover most TM use cases and enable both experienced practitioners and beginners to use TMs.        
        One example of this principle is that GT provides hyperparameter search and Cross-Validation (CV) out of the box.\\
    }
    
    \item {
        \textbf{Hardware Agnostic} \\
        The framework should be hardware-agnostic. However, where more advanced capabilities are available, it should gracefully utilize those instead.
        For instance, if a CPU supports the AVX2 instruction set, it should automatically upgrade its backend from a pure C++ version to one that
        takes advantage of AVX2. Conversely, it should also be able to fall back to a pure Python version if no C++ backend is available. The same goes for ARM CPUs that do not support NEON intrinsics.\\
    }
    
    \item{
        \textbf{Python Orchestration for training} \\
         Python should be used to orchestrate a computational graph executed in a
         high-performance environment for training a TM. This allows us to leverage a fast backend while still keeping a flexible Python interface.\\
    }

    \item{
        \textbf{Seperation between Training and Inference} \\
        GT is built around what typical TM tasks consist of.
        First, train a TM to obtain clauses and weights, then run inference on the TM. 
        Therefore, GT should clearly distinguish between these two modes, each optimized for its purpose.
        For instance, GT supports exporting a trained TM as a standalone C file for easy integration with 
        embedded systems or high throughput inference.
    }
\end{itemize}

\subsection{Project Openness and Development}
GT is released under the MIT license. The entire source code, including examples, can be
found at: \\
\url{https://github.com/ooki/green_tsetlin}. 
An extensive suite of automated tests covers the codebase, ensuring that the released versions are as stable as possible. 

\section{Why Green Tsetlin?}

The general difference between the various existing open-source frameworks and GT 
can be found in Table \ref{tab:comparison}. To summarize, GT provides more wrapping around the journey from training to production.
And is, as time-of-writing, the only implementation of SparseTsetlin \cite{ostby2024sparse}, allowing it to handle previously
unseen data scales. For instance, training on 8M documents (Amazon Reviews) with a vocabulary of 5M only requires a TM of 1 MiB as opposed
to 18.6 GiB for a dense TM. The TMU SparseClause bank implementation requires an initial underlying dense state representation that gradually
is sparsified.

\begin{table}[]
\begin{tabular}{l|l|l|l|l|l}
\hline
Framework   /    \#Clauses:        & 10   & 100  & 1000 & 2000  & 5000  \\ \hline
TMU (CPU)                         & 29.6 & 33 & 80 & 128 & 266 \\ \hline
TMU (GPU)                         & 87.8 & 62 & 70 & 75  & 95  \\ \hline
pyTsetlinMachine                  & 4.5  & 9  & 47 & 84  & 197 \\ \hline
pyTsetlinMachineParallel (jobs=4) & 1.1  & \textbf{2}  & 17 & 47  & 149 \\ \hline
PyCoalescedTsetlinMachineCUDA     & 14.8 & 26 & 30 & 36  & 88  \\ \hline
GT (jobs = 1)                     & \textbf{0.3}  & \textbf{2}  & 18 & 36  & 90  \\ \hline
GT (jobs = 4)                     & 4.0  & 5  & \textbf{11} & \textbf{18}  & \textbf{31}  \\ \hline
\end{tabular}
\caption{Time in seconds needed to train 5 epochs on MNIST, where the test set is evaluated after each epoch.
With an increasing number of clauses. Hardware used: i9-12900K CPU, GPU: GeForce RTX 3090.
All frameworks have roughly the same accuracy, so they are omitted for brevity.}
\label{tab:performance}
\end{table}

Table \ref{tab:performance} shows a runtime comparison, demonstrating that GT is faster than most existing frameworks. Note that
not all the frameworks implement Coaleased TM \cite{glimsdal2021coalesced} (pyTsetlinMachine, pyTsetlinMachineParallel) and Constrained Clause Sizes \cite{abeyrathna2023building} (pyTsetlinMachineParallel, PyCoalescedTsetlinMachineCUDA). However, it is not intended as a comprehensive performance test, as the various frameworks excel in different settings. The CUDA based frameworks will scale better
for bigger and dense datasets.

\begin{table*}[ht]
\begin{tabular}{|c|c|c|c|c|c|c|l|l|l|}
\hline
                              & Dense & Coaleased TM & Sparse & Multithreading & CUDA        & Inference   & AVX2        & NEON        & Export      \\ \hline
pyTsetlinMachine              & \greencheck &              &              &                &             &             &             &             &             \\ \hline
pyTsetlinMachineParallel      & \greencheck &              &              & \greencheck    &             &             &             &             &             \\ \hline
PyCoalescedTsetlinMachineCUDA & \greencheck & \greencheck  &              &                & \greencheck &             &             &             &             \\ \hline
TMU                           & \greencheck & \greencheck  &              &                & \greencheck &             &             &             &             \\ \hline
Green Tsetlin                 & \greencheck & \greencheck  & \greencheck  & \greencheck    &             & \greencheck & \greencheck & \greencheck & \greencheck \\ \hline
\end{tabular}
\caption{Dense: Running on a fully specified TM space.
Coaleased TM: Supports clause sharing between the classes \cite{glimsdal2021coalesced}.
Sparse: Supports Sparse Data and Sparse TM (where the TM state is sparse).
Multithreading: Supports multithreading. CUDA: Supports CUDA Acceleration.
Explainable: Explainable tooling is directly supported (such as \cite{blakely2020closedform}).
AVX2: Utilize AVX2 intrinsics if available.
NEON: Utilize NEON intrinsics if available (found in ARM CPUs such as Apple Sillicon and Rasberry PI).
Export: Export a trained TM as a standalone program or an inference module ready for production tasks.}
\label{tab:comparison}
\end{table*}

\section{Overview}
\begin{figure*}[!t]
\centering
\includegraphics[width=7in]{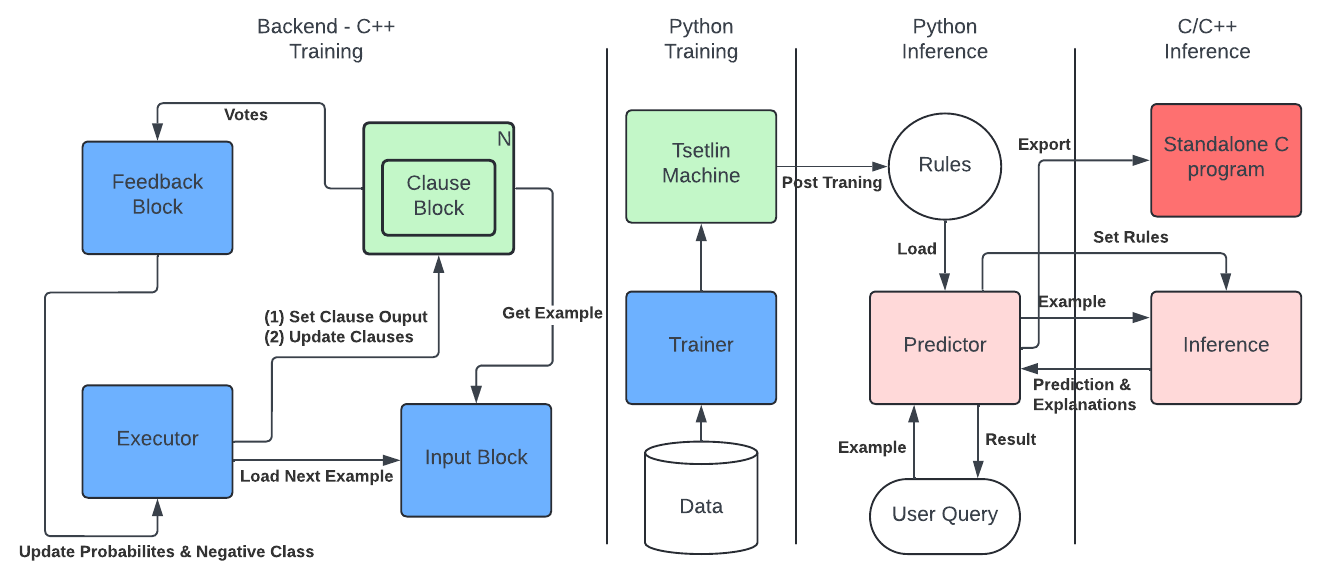}
\caption{Outline of the various components of GT. The Python box of the corresponding color
orchestrates the backend blocks (in C++). For instance, the responsibility of creating \& adding
the ClauseBlocks to the computation graph falls to the Python Tsetlin Machine class (green color).
The red box is the exportation of the TM rules into its own standalone C program.
}
\label{fig:traning_overview}
\end{figure*}

An overview of the flow of GT is shown in Figure \ref{fig:traning_overview}.
The computational graph is composed of four block types:
\begin{itemize}
    \item{
        ClauseBlock: This is the basic building block of GT; it contains the clauses and weights in addition to the logic for evaluating and updating clauses. In a multithreaded scenario, each Clause Block is run in a separate thread for easy parallel execution.
    }
    \item{
        Input Block: Contains the training data and feeds the clause blocks with examples.
    }

    \item{
        Feedback Block: Collects the votes from each of the Clause Blocks and calculates the 
        update probability for both the positive and negative classes. 
    }

    \item{
        Executor: Facilitates the training loop either in a parallel execution or a single thread execution.
    }
\end{itemize}

\subsection{The training loop}
(1) The input block prepares an example.
(2) The executor tells each clauseblock to load an example from the input block and calculate clause outputs and class votes.
(3) The feedback block registers votes from each clause block and calculates an update probability alongside a negative target.
(4) Finally, the executor provides each clause block with a positive and negative target alongside their update probabilities. The clause blocks then
updates all their clauses and weights.

Using the Python interface, this orchestration is reduced to: 
\begin{minted}
[
frame=lines,
framesep=2mm,
baselinestretch=1.0,
fontsize=\footnotesize,
]
{python}
import green_tsetlin as gt

tm = gt.TsetlinMachine(n_literals=6,
                       n_clauses=8,
                       n_classes=2,
                       s=1.9,
                       threshold=11,
                       n_literal_budget=3)
trainer = gt.Trainer(tm, n_epochs=3,
                         seed=42,
                         n_jobs=-1)
trainer.set_train_data(train_x, train_y)
trainer.set_eval_data(test_x, test_y)
trainer.train()        
\end{minted}

\subsection{Inference}
The predictor class is used to perform predictions on a trained TM. A predictor 
is created by compiling a TM into a set of weighted rules.
The inference processes, complete with explanations that operate on both the literal and feature levels, can then either be carried out by the predictor object itself or exported
as a C file for further compilation into a standalone program.

\begin{minted}
[
frame=lines,
framesep=2mm,
baselinestretch=1.2,
fontsize=\footnotesize,
]
{python}
// cont example
predictor = tm.get_predictor(explanation="literals")

// Using the predictor
y_hat, explanation = predictor.predict_and_explain(x)

// Or export as a C header file
predictor.export_as_program("inference_tm.h")
\end{minted}

\section{Conclusion \& Future Work}
This paper introduces Green Tsetlin, a robust and lightweight Tsetlin Machine (TM) framework designed with a production-focused approach that offers efficiency beyond accelerated TM training. This well-documented framework incorporates essential features required for developing high-quality machine learning products.

Green Tsetlin is versatile and capable of operating in both dense and sparse modes. As we look to the future, we aim to enhance its capabilities by integrating CUDA support and addressing additional requirements for production environments as they are published.

Our vision is for Green Tsetlin to serve as a user-friendly and accessible pathway for new and experienced TM users, enabling them to create and deploy powerful machine-learning solutions easily.

\section*{Acknowledgment}
We want to thank Forsta Labs for their support with this work.
Also, big thanks to everyone who has given us valuable feedback on the development of GT.

\bibliographystyle{IEEEtran}
\bibliography{bibliography}

\begin{thebibliography}{1}
\providecommand{\url}[1]{#1}
\csname url@samestyle\endcsname
\providecommand{\newblock}{\relax}
\providecommand{\bibinfo}[2]{#2}
\providecommand{\BIBentrySTDinterwordspacing}{\spaceskip=0pt\relax}
\providecommand{\BIBentryALTinterwordstretchfactor}{4}
\providecommand{\BIBentryALTinterwordspacing}{\spaceskip=\fontdimen2\font plus
\BIBentryALTinterwordstretchfactor\fontdimen3\font minus \fontdimen4\font\relax}
\providecommand{\BIBforeignlanguage}[2]{{%
\expandafter\ifx\csname l@#1\endcsname\relax
\typeout{** WARNING: IEEEtran.bst: No hyphenation pattern has been}%
\typeout{** loaded for the language `#1'. Using the pattern for}%
\typeout{** the default language instead.}%
\else
\language=\csname l@#1\endcsname
\fi
#2}}
\providecommand{\BIBdecl}{\relax}
\BIBdecl

\bibitem{granmo2021tsetlin}
O.-C. Granmo, ``The tsetlin machine -- a game theoretic bandit driven approach to optimal pattern recognition with propositional logic,'' 2021.

\bibitem{glimsdal2021coalesced}
S.~Glimsdal and O.-C. Granmo, ``Coalesced multi-output tsetlin machines with clause sharing,'' 2021.

\bibitem{ostby2024sparse}
S.~Østby, T.~M. Brambo, and S.~Glimsdal, ``The sparse tsetlin machine: Sparse representation with active literals,'' 2024.

\bibitem{abeyrathna2023building}
K.~D. Abeyrathna, A.~A.~O. Abouzeid, B.~Bhattarai, C.~Giri, S.~Glimsdal, O.-C. Granmo, L.~Jiao, R.~Saha, J.~Sharma, S.~A. Tunheim, and X.~Zhang, ``Building concise logical patterns by constraining tsetlin machine clause size,'' 2023.

\bibitem{blakely2020closedform}
C.~D. Blakely and O.-C. Granmo, ``Closed-form expressions for global and local interpretation of tsetlin machines with applications to explaining high-dimensional data,'' 2020.

\end{thebibliography}

\end{document}